\documentclass[journal]{IEEEtran}
\usepackage{xcolor}

\usepackage{dsfont}
\usepackage[utf8]{inputenc} 
\usepackage[T1]{fontenc}    
\usepackage{hyperref}       
\usepackage{url}            
\usepackage{booktabs}       
\usepackage{amsfonts}       
\usepackage{nicefrac}       
\usepackage{microtype}      

\usepackage{csvsimple}
\usepackage{caption}
\usepackage{graphicx}
\graphicspath{ {./Figures/} }
\usepackage{caption}
\usepackage{subcaption}
\usepackage{amsmath}
\DeclareMathOperator*{\argmax}{argmax} 

\usepackage[export]{adjustbox}

\usepackage{amsmath} 
\usepackage{graphicx}
\usepackage{caption}
\usepackage{booktabs}
\usepackage{amssymb}
\usepackage{bm}
\usepackage[ruled,vlined]{algorithm2e}
\usepackage{caption}
\usepackage[noend]{algpseudocode}
\usepackage{url}

\usepackage[nameinlink,capitalise]{cleveref}
\usepackage{multirow}
\usepackage{xcolor}
\usepackage{longtable}
\usepackage{graphicx}
\usepackage{balance}
\usepackage{amsmath}
\usepackage{tabularx,booktabs}
\newcolumntype{C}{>{\centering\arraybackslash}X} 
\title{Texture based Prototypical Network for Few-Shot Semantic Segmentation of Forest Cover: Generalizing for Different Geographical Regions }
 
\author{Gokul P,
       and Ujjwal~Verma ,~\IEEEmembership{Senior Member,~IEEE}
\IEEEcompsocitemizethanks{\IEEEcompsocthanksitem Gokul P is  with the Department of Electronics and Communication Engineering, Manipal Institute of Technology, Manipal, Manipal Academy of Higher Education, India.\protect\\
\IEEEcompsocthanksitem Ujjwal Verma is with Department of Electronics and Communication Engineering, Manipal Institute of Technology Bengaluru, Manipal Academy of Higher Education, India. Email: ujjwal.verma@manipal.edu}
}
        
\begin{document}
\maketitle

\begin{abstract}
 Forest plays a vital role in reducing greenhouse gas emissions and mitigating climate change besides maintaining the world's biodiversity. The existing satellite-based forest monitoring system utilizes supervised learning approaches that are limited to a particular region and depend on manually annotated data to identify forest. This work envisages forest identification as a few-shot semantic segmentation task to achieve generalization across different geographical regions. The proposed few-shot segmentation approach incorporates a texture attention module in the prototypical network to highlight the texture features of the forest. Indeed, the forest exhibits a characteristic texture different from other classes, such as road, water, etc.  In this work, the proposed approach is trained for identifying tropical forests of South Asia and adapted to determine the temperate forest of Central Europe with the help of a few (one image for 1-shot) manually annotated support images of the temperate forest.  An IoU of 0.62 for forest class (1-way 1-shot) was obtained using the proposed method, which is significantly higher (0.46 for PANet) than the existing few-shot semantic segmentation approach. This result demonstrates that the proposed approach can generalize across geographical regions for forest identification, creating an opportunity to develop a global forest cover identification tool.

\end{abstract}

\section{Introduction}
Forest plays a vital role in regulating the ecosystem, protecting biodiversity, and supporting sustainable growth. It does also act as a  significant carbon sink absorbing one-third of the $CO_{2}$ released from burning fossil fuels. Due to the increasing demand for agricultural farmland, timber, mineral, and other resources, there has been a continuous decrease in forest cover across the globe. The world has lost around 178 million \textit{hectares} of forest since 1990, with Africa recording the largest annual rate of net forest loss  \cite{FAO}. Deforestation severely affects the carbon cycle, making it a significant focus area for climate change studies. Besides, deforestation also results in biodiversity loss, adversely affecting the surrounding ecosystem.

The satellite-based forest monitoring system provides a cost-effective approach for studying deforestation. This system compares the forest cover of a particular region with the earlier forest cover of the same region to locate the area of deforestation. Identifying forests in satellite images acquired over a period of time is beneficial for detecting deforestation in remote and inaccessible areas.  For instance, the Brazilian government's PRODES and DETER program uses Landsat and MODIS data to calculate the deforestation region in Brazil's legal amazon. Despite the success of these approaches in identifying the areas of deforestation, these methods are limited to Brazil's legal amazon rainforest and have limited success in identifying small-scale deforestation. Besides, these approaches involve manual analysis, which can be a time-consuming task \cite{isprs-annalsOrtega}. In addition, change detection techniques have also been utilized to identify regions of deforestation \cite{TEWKESBURY20151}. However, these approaches' study area is limited to a particular region such as the Brazilian Amazon \cite{Bam2020change}.

Forest identification in satellite images is a crucial step in a satellite-based forest monitoring system. In this work, forest identification is formulated as a few-shot semantic segmentation task. In a supervised learning approach, the train and test distribution are assumed to be identical, which would limit their application to the geographical region represented in the training set. In contrast, the few-shot learning can be utilized to identify the forest of an unseen region, which is \textit{not} present in the training set, with the help of a few labelled images of the unseen region. Unlike the supervised learning approach, the few-shot method can be utilized to identify forests, especially for regions where only a few labelled images (support set) are available \cite{russwurm2020meta}. Therefore, this approach can be used to create a tool for identifying forests across all regions with a limited amount of manually annotated images  (one image for 1-shot). Despite the recent advances in few-shot learning, few authors have utilized this approach for forest identification. The authors in \cite{russwurm2020meta} studied the performance of few-shot learning (model agnostic meta-learning (MAML)) algorithm for land cover segmentation on high-resolution satellite images of South Asia. In \cite{russwurm2020meta}, the original images were divided into non-overlapping sub-images, and a few sub-images were used as support set to perform the segmentation on other sub-images.  This approach permitted the authors to study the model's performance in identifying the entire region's forest using a small number of sub-images of the \textit{same} geographical region.

In comparison, our work focuses on evaluating the few-shot semantic segmentation method's performance in identifying the forest of a particular geographical region using the model trained on a completely \textit{different} geographical region. This work proposes a prototypical  few- shot semantic segmentation network that leverages a texture attention module. The tree canopy of the forest exhibits a characteristic texture different from other background classes, such as roads, barren lands etc.  The model is trained on DeepGlobe land cover segmentation dataset \cite{DeepGlobe18}, which contains satellite images from India, Indonesia, and Thailand. This trained model is then evaluated on the LandCover AI dataset \cite{landcovera}, which includes images from Central Europe. This formulation of few-shot semantic segmentation allows us to develop a forest identification method, which would not be restricted by the geographical region. In this work, forest refers to the region covered with trees standing in proximity with single trees and orchards being ignored as defined in \cite{landcovera}.

This work's main contributions are: 1) Re-formulation of few-shot semantic segmentation task for identifying forest across different geographical region: Training the model on images of tropical forests (South Asia), while testing its performance on temperate forests (Central Europe),  2) Development of texture based prototypical few-shot segmentation architecture for forest identification.  The rest of the paper is organized as follows: Section \ref{met} presents the proposed few-shot semantic segmentation approach utilized for forest identification, and Section \ref{Sec:Results} shows the result obtained.

\begin{figure*}[!htb]
\begin{center}
\includegraphics[width=1\linewidth]{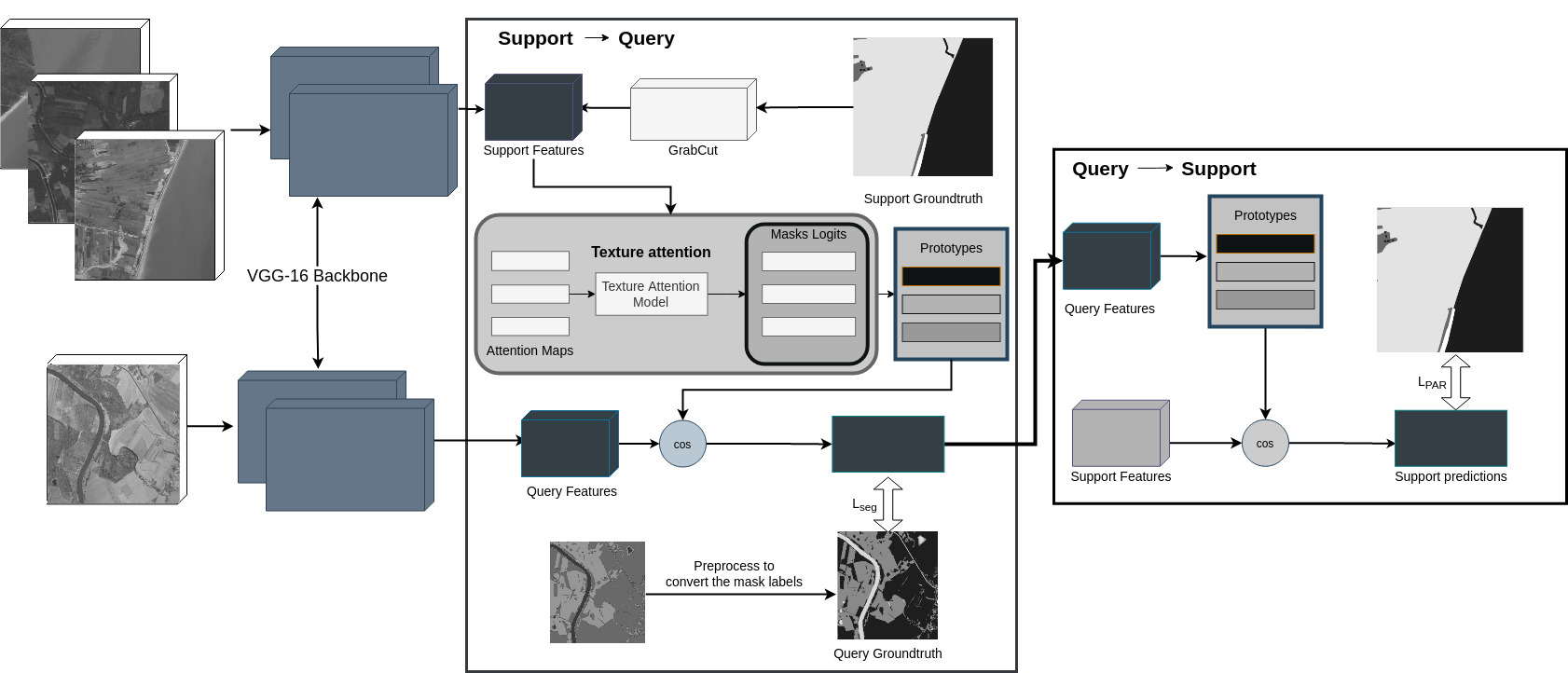}
\caption{Overview of the proposed texture-based prototypical network for identifying forests across geographical regions.}\label{Fig:Overview}
\end{center}

\end{figure*}

\section{Method}
\label{met}
\subsection{Problem Formulation}
This work focuses on identifying areas with forest cover across geographical regions with limited annotated images. The cross-geography generalization is achieved by formulating the forest identification as a few-shot segmentation task. In a few shot semantic segmentation task, images from two non overlapping sets of classes, $C_{train}$ and $C_{test}$ are provided such that training dataset, $D_{train}$ consists of images of $C_{train}$ classes and test dataset $D_{test}$ is made from $C_{test}$ \cite{wang2019panet}. In this work, the $D_{train}$ contains images from South East Asia \cite{DeepGlobe18} while $D_{test}$ contains images from Central Europe \cite{landcovera}. This evaluation paradigm focuses on cross-geography generalization by utilizing tropical forest images of South Asia for training the model while testing the model on temperate forest images of central Europe.

As in the few shot semantic segmentation framework \cite{wang2019panet} , the train/test set is divided into several episodes, with each episode containing a set of support image with annotation and the query image. A C-way K-shot semantic segmentation implies that support set $S$  contains K image pairs (Image, Mask) with pixels from C different classes per episode.  The goal is to segment the area of the unseen class $C_{test}$ from each query image given $K$ samples from the support set $S$.

\subsubsection{Dataset and Pre-Processing}
\label{data_aug}

The dataset from the DeepGlobe challenge \cite{DeepGlobe18} contains high-resolution optical satellite images from  South Asian countries, including India, Thailand, and Indonesia. The dataset contains 803 images of 2,448 x 2,448 resolution. A mask is provided for each image which contains seven different classes: Urban, Agriculture, Rangeland, Forest, Water, Barren and also an Unknown class.  In this work, the images are tiled to create sub-images of 612 x 612 pixels. This tiling results in 12,800 sub-images that form the training set, $D_{train}$. Moreover, this work focuses on forest identification; therefore the following classes are merged for 1-way task: Urban, Agriculture, Water, Rangeland, Barren and Unknown. 

On the other hand, the test dataset was adapted from the LandCover AI dataset \cite{landcovera} which consists of 41 RGB images acquired over central Europe (Poland). This dataset contains four classes: Building, Woodlands (trees standing in close proximity), Water, and Background (unknown). 
The input images in train set were resized to (128 x 128) pixels and augmented by randomly flipping the image/mask horizontally. According to the standard practice for few-shot semantic segmentation, this study fixed the number of query images as one. For a 1-way task, the classes in $D_{train}$ are Forest and Background, while for 2-way task, the classes are Forest, Water and Background. The background class contains all other classes present in the dataset apart from forest (for 1-way) and forest, water (for 2-way).Since we are using two different datasets with different labeling policies, we have pre-processed the ground truth masks to ensure that pixel-level labels are represented uniformly in train and test split.

\subsection{Method Overview}

The proposed method first extracts the prototypes from the support images (Section \ref{SubsubSec:SuppProto}). Unlike the existing methods, this work emphasizes the textures features while computing the support prototypes. Indeed, the pixels corresponding to the forest exhibit characteristic textures. Subsequently, the query prototype is extracted from the query image. Finally, segmentation mask for the query image is computed using metric learning between the prototype and support features. In addition, prototype alignment regularization \cite{wang2019panet} is also implemented which learns a better representation from the support images (Figure\ref{Fig:Overview}) . 

\subsubsection{Texture based Support Image Prototype}
\label{SubsubSec:SuppProto}
Given the set of support images $S$, VGG-16 \cite{simonyan2014very} as feature extractor is utilized to compute the features from the support images. The five convolutional block of VGG-16 with maxpool4 layer set to 1 and dilated convolution (dilation = 2) is utilized in this work. Let us represent these features as $F_{k}$ for the $k^{th}, k = 1,2,...,K$ image in the support set. The feature vector $F_{k}$ are resized to the same size as the input image using bilinear interpolation \cite{zhang2020sgone}. The texture attention module then enhances these support features to highlight the characteristic textures of forests.

\paragraph{Texture attention module}: The texture features are extracted as an activation of a Gabor Filter based CNN \cite{alekseev2019gabornet}. The filters for the first layer of this CNN are constrained to fit Gabor function. The filter parameters are initialized from the Gabor filter bank, and then updated by backpropagation. Let us represent $T_{k}$ as the texture features extracted for a given image $k$ using the Gabor CNN. Note that this feature vector is resized to the same size as the input image using bilinear interpolation \cite{zhang2020sgone}. 

The texture features of only the pixels corresponding to the forest are utilized as attention maps. These forest pixels are identified using GrabCut \cite{GrabCut} based foreground segmentation approach. GrabCut uses an improved iterative Graph Cut based technique coupled with border matting. 
The method involved iterating over the \textit{support} images while using the support image ground truth mask as the trimap reference keeping the interaction rectangle to be the size of the image. This is then run through a max-flow method to determine the min-cut, which once converges, separates the background from the foreground (forest). Let us represent the binary mask generated by Grab Cut as $G_{k} \in {0,1}$ representing the foreground (forest) regions. Note that using a foreground segmentation approach rather than a ground truth segmentation mask eliminates the label noise that might be present in the manually annotated data.

Subsequently, the texture features for only forest pixels are computed. Specifically, the final texture features from the support image is computed as

\begin{equation}
    \hat{T}_{k} = T_{k}(x,y) G_{k}(x,y)
\end{equation}
where $T_{k}$ is the texture features and $G_{k}$ is the binary mask identifying foreground (forest) and background.

The support prototype for forest class is computed using masked average pooling 
\begin{equation}
    p_{forest} = \frac{1}{K} \sum_{k} \frac{\sum_{(x,y)}\hat{T}_{k}(x,y) F_{k}(x,y)}{\sum_{(x,y)}G_{k}(x,y)}
\end{equation}

The support prototype for the background class $p_{bg}$ is computed from the masked average pooling of the VGG-16 features as follows:
 \begin{equation}
     p_{bg} = \frac{1}{K}\sum_{k} \frac{\sum_{(x,y)} G^{'}_{k}(x,y)F_{k}(x,y) }{\sum_{(x,y)} G^{'}_{k}(x,y)}
    \end{equation}
 where $G^{'}_{k}(x,y) = ! G_{k}(x,y)$, $!$ is the logical NOT operator.

\subsubsection{Metric Learning: Support - Query}
\label{SubSecMetricLearning}
For each spatial location in the query image, the optimal class is computed by comparing the distance between support prototype features and features computed from query images using the approach proposed in \cite{wang2019panet}. A brief summary of the metric learning approach is presented here. More details can be found in \cite{wang2019panet}. The features from each query image is computed using VGG-16 as a feature extractor. Let us represent query features as $F_{q}$. The distance $d$ between query features and support prototype can be defined as
\begin{equation}
    dist_{j} = d(p_{j},F_{q})
\end{equation}
where $p_{j} $ is either $p_{forest}$ or $p_{bg}$. In this work, cosine distance is considered. The segmentation mask $ M_{q}(x,y)$ can be then be computed as \cite{wang2019panet}:
\begin{equation}
    M_{q}(x,y) = \argmax_j \frac{\exp{(-\alpha ~ dist_{j}})}{\sum_{p_{j}} \exp{(-\alpha ~ dist_{j}})}
\end{equation}
where $\alpha$ is a constant. 
The segmentation loss $L_{seg}$ is then computed as a cross entropy loss between the segmentation mask $ M_{q}$ and the ground truth segmentation mask. 
\subsubsection{Regularization: Query - Support}
A prototype alignment regularization proposed in \cite{wang2019panet} is also utilized in this work. In this approach, few shot learning is performed in the reverse direction, i.e., given the query image and its predicted mask, compute the segmentation mask for support image. Specifically, features from query and support images are computed using VGG-16 backbone, and masked average pooling is utilized to compute the query and support prototypes. Note that the texture features are not computed in this regularization step. The cosine similarity-based approach (Section \ref{SubSecMetricLearning}) is then used to compare the query and support prototype and find predicted mask for the support image. Finally, the regularization loss ($L_{PAR}$) is defined as the cross-entropy loss between the predicted mask for support image and its ground truth segmentation mask. The total loss used for training the model is a weighted sum of segmentation loss ($L_{seg}$) and the regularization loss ($L_{PAR}$).

\section{Results and Discussion}
\label{Sec:Results}

The proposed model was trained with a Stochastic Gradient Descent optimizer with a learning rate of 1e-3 and a momentum of 0.9 for 30,000 iterations. The model's performance was evaluated by comparing the manually annotated ground truth class labels with the predicted class labels for each pixel in the image and computing mean Intersection over Union (mIoU). 

First, the contribution of each module (viz. Support to Query (S2Q), Query to Support (Q2S) and Texture Attention (TA)) was evaluated. Tables \ref{tab:mIouNEW} and \ref{tab:ForestNew} shows the overall mIoU and forest class IoU for the following scenarios: Only Support to Query (S2Q), Support to Query with Texture Attention (S2Q-TA), Support to Query and Query to Support with Texture Attention (S2Q + Q2S - TA), and Support to Query with Texture Attention and Query to Support with Texture Attention (S2Q - TA + Q2S -TA). It can be observed that the inclusion of the texture attention module in support to query module results in a more accurate segmentation than support to query module without textures. Interestingly, the support to query module with texture performs better than PANet (S2Q + Q2S). Moreover, the inclusion of texture information in both support to query and query to support performs competitively when texture information is included only in support to query module. Therefore, the proposed method includes texture information only in the support to query module.

The proposed method is compared with the existing few-shot semantic segmentation methods (PANet \cite{wang2019panet}, PFENet \cite{PFENetTian2020} and ASGNet \cite{ASGNet2021}). As discussed earlier, training dataset contains images from South Asian countries, while the test dataset contains images from central Europe (Section \ref{met}, A).  Table \ref{tab:mIouNEW} shows the mIoU obtained for 1-way and 2-way tasks. It can be observed that the proposed approach outperforms the existing few shot segmentation methods. Notably, the proposed method achieves a higher mIoU (0.47 for 2-way 5-shot) as compared to PANet \cite{wang2019panet} (0.40 for 2-way 5-shot).  Besides, the proposed method achieves an IoU of 0.69 for the forest class on 2-way 5-shot task, which is significantly higher than the existing methods (Table \ref{tab:ForestNew}). Note that PANet uses VGG-16 features for computing prototypes and does not include texture features. The higher mIoU demonstrates the effectiveness of the proposed texture attention module.
Indeed, forest exhibits characteristics texture feature which is distinct from other background classes, such as building, water, roads etc. The higher IoU demonstrates that the proposed approach can transfer the representation learned of South Asian forests to identify Central Europe's forests with the help of only limited support images of Central Europe. It can be seen that the model can identify temperate forests with the help of only \textit{one} labeled image, even though it was trained for identifying tropical forests. 

\begin{figure}

     \centering
     \begin{adjustbox}{minipage=\linewidth,scale=1}
     \begin{subfigure}[b]{0.23\textwidth}
         \centering
         \includegraphics[width=\textwidth]{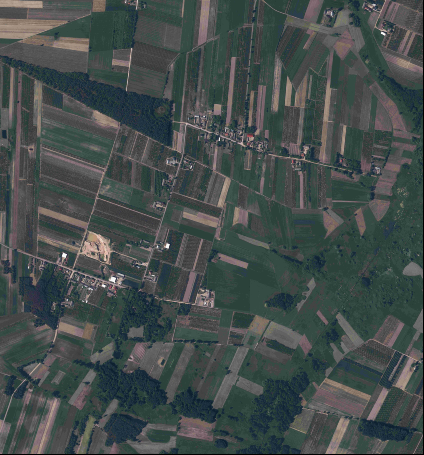}
         \label{fig:y equals x}
     \end{subfigure}
      \hfill
      \begin{subfigure}[b]{0.24\textwidth}
         \centering
         \includegraphics[width=\textwidth]{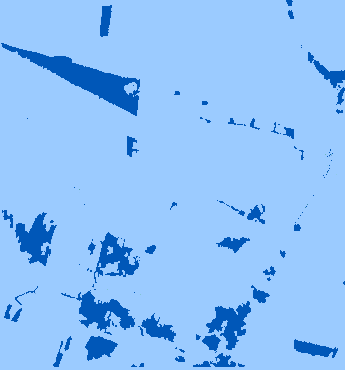} 
         \label{fig:five over x}
     \end{subfigure}
     \hfill
     \begin{subfigure}[b]{0.24\textwidth}
         \centering
         \includegraphics[width=\textwidth]{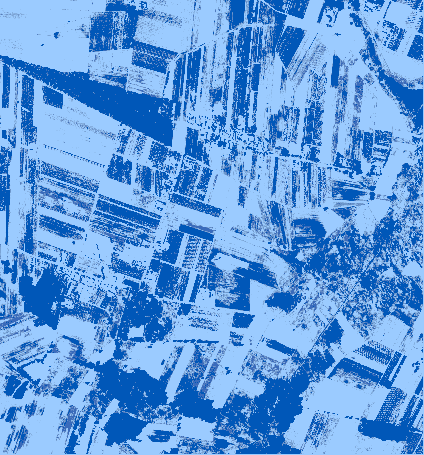}
         \label{fig:three sin x}
     \end{subfigure}
     \hfill
     \begin{subfigure}[b]{0.24\textwidth}
         \centering
         \includegraphics[width=\textwidth]{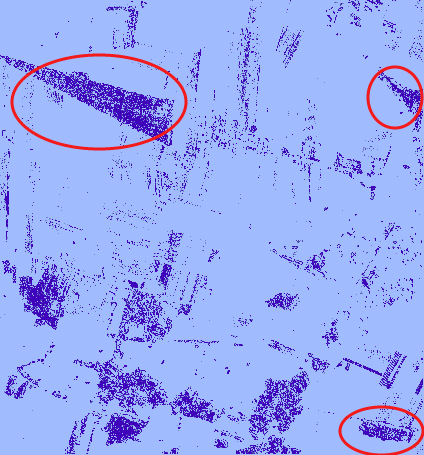}
         \label{fig:five over x}
     \end{subfigure}
     \hfill

     \begin{subfigure}[b]{0.24\textwidth}
         \centering
         \includegraphics[width=\textwidth]{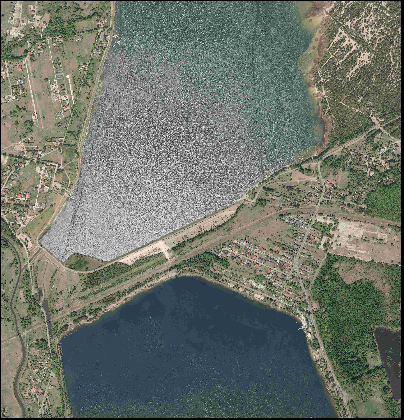}
         \label{fig:y equals x}
     \end{subfigure}
     \hfill
     \begin{subfigure}[b]{0.24\textwidth}
         \centering
         \includegraphics[width=\textwidth]{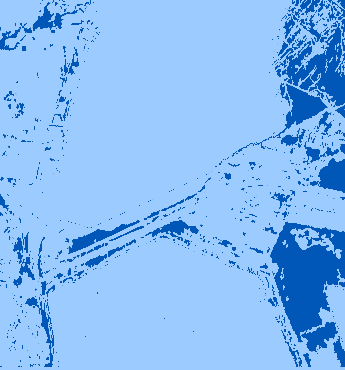}
         \label{fig:five over x}
     \end{subfigure}
     \hfill
     \begin{subfigure}[b]{0.24\textwidth}
         \centering
         \includegraphics[width=\textwidth]{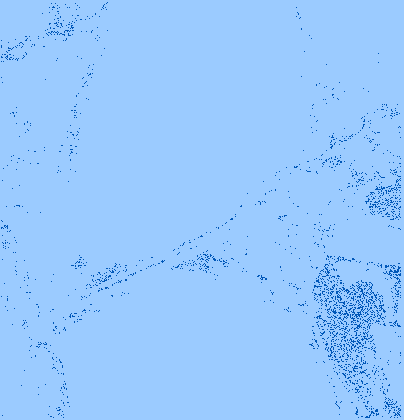}
         \label{fig:three sin x}
     \end{subfigure}
     \hfill
     \begin{subfigure}[b]{0.24\textwidth}
         \centering
         \includegraphics[width=\textwidth]{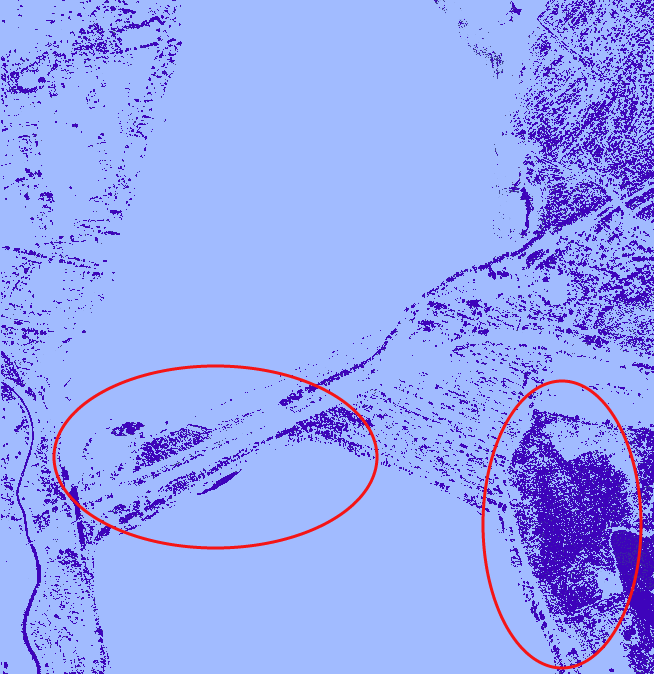}
         \label{fig:five over x}
     \end{subfigure}

      \begin{subfigure}[b]{0.24\textwidth}
         \centering
         \includegraphics[width=\textwidth]{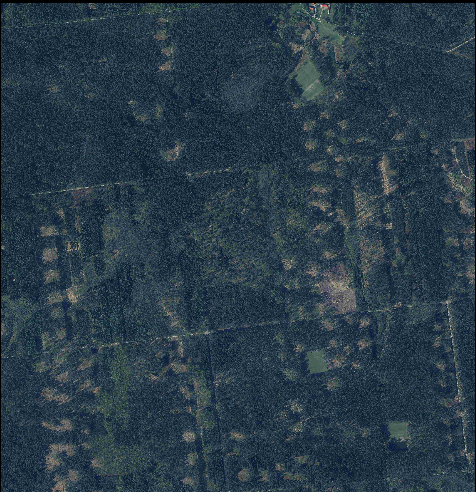}
         \label{fig:y equals x}
     \end{subfigure}
     \hfill
     \begin{subfigure}[b]{0.24\textwidth}
         \centering
         \includegraphics[width=\textwidth]{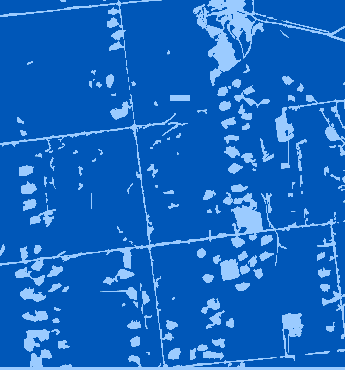}
         \label{fig:five over x}
     \end{subfigure}
     \hfill
     \begin{subfigure}[b]{0.24\textwidth}
         \centering
         \includegraphics[width=\textwidth]{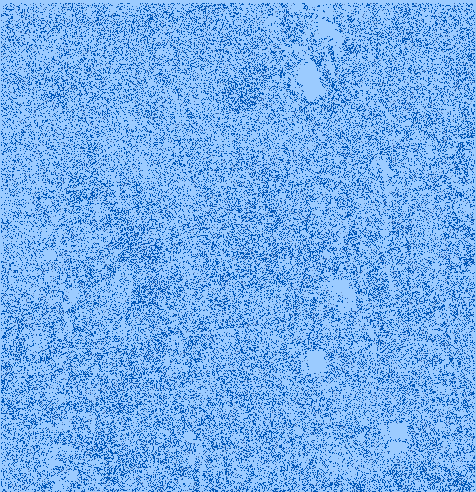}
         \label{fig:three sin x}
     \end{subfigure}
     \hfill
     \begin{subfigure}[b]{0.24\textwidth}
         \centering
         \includegraphics[width=\textwidth]{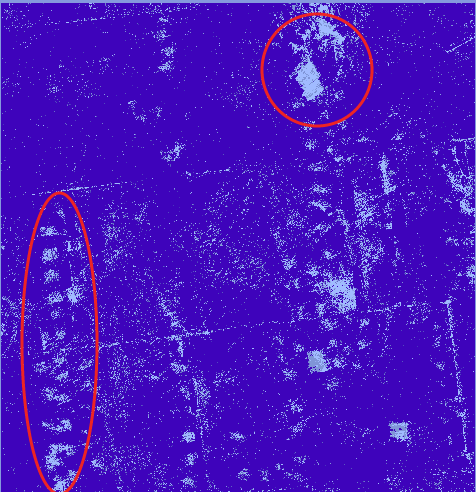}
         \label{fig:five over x}
     \end{subfigure}

      \begin{subfigure}[b]{0.24\textwidth}
         \centering
         \includegraphics[width=\textwidth]{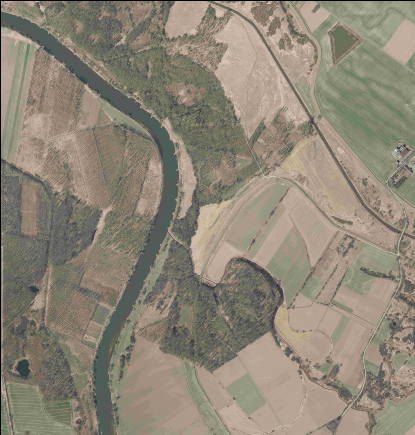}
         \caption{Original}
         \label{fig:y equals x}
     \end{subfigure}
     \hfill
     \begin{subfigure}[b]{0.24\textwidth}
         \centering
         \includegraphics[width=\textwidth]{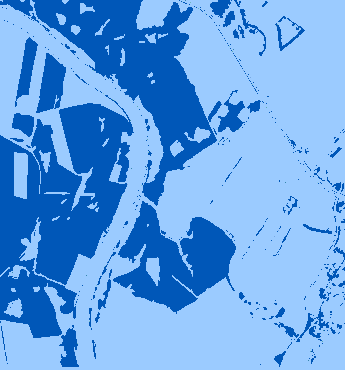}
         \caption{Ground Truth}
         \label{fig:five over x}
     \end{subfigure}
     \hfill
     \begin{subfigure}[b]{0.24\textwidth}
         \centering
         \includegraphics[width=\textwidth]{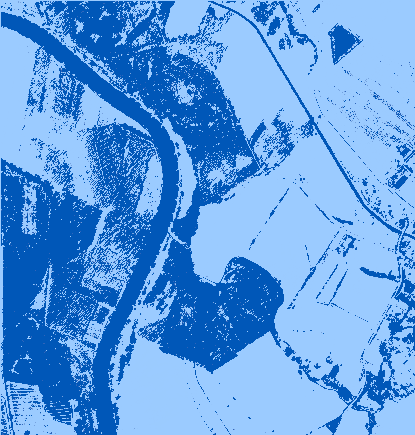}
         \caption{PANet}
         \label{fig:three sin x}
     \end{subfigure}
     \hfill
     \begin{subfigure}[b]{0.24\textwidth}
         \centering
         \includegraphics[width=\textwidth]{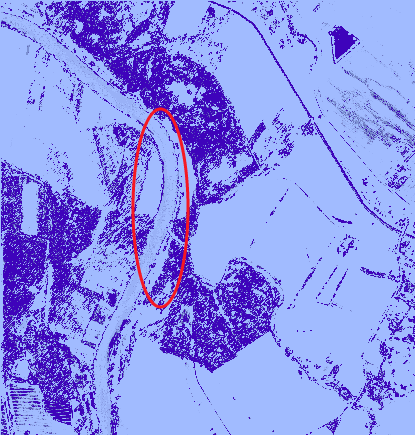}
         \caption{Ours}
         \label{fig:five over x}
     \end{subfigure}
     
 \end{adjustbox}   
     
        \caption{Comparing the performance of the proposed method with PANet \cite{wang2019panet} in identifying temperate forest of Central Europe. The first and second column represent the original image and its corresponding ground truth mask, while the third and fourth column shows the results obtained using PANet and the proposed method respectively for 2-way 5-shot task. Dark blue represents forest class while light blue represents other classes.  }
        \label{fig:three graphs}
\label{Fig:results}
 
\end{figure}

Figure \ref{Fig:results} compares the result obtained using the proposed method with PANet \cite{wang2019panet} on few images for 2-way 5-shot task. It can be observed that region representing the forest are identified more accurately as compared to PANet. Notably, the proposed method can distinguish between cropland and forest regions as shown in Figure \ref{Fig:results} (First Row, red circles). Note that PANet \cite{wang2019panet} identifies the cropland as a forest region apparently due to similar color information.  Besides, the proposed method can differentiate between other classes (water, open lands) and forest as compared to PANet (Figures \ref{Fig:results}, second and third row). Interestingly, PANet identifies the river as a forest region due to the similar color information (Figure \ref{Fig:results}, fourth row). However, the proposed method identifies only the forest region. These results demonstrate the effectiveness of the texture attention module in discriminating between the forest and other classes.

To further highlight the importance of texture features for forest identification,  the proposed texture attention module was integrated in two existing few-shot segmentation methods (PFENet \cite{PFENetTian2020} and ASGNet \cite{ASGNet2021}). For ASGNet, the proposed texture attention map was multiplied with the features from the “expand block” for single prototype learning.  For PFENet, the texture attention map was multiplied with the features obtained from the masked global pooling module. The mIoU obtained using these modified models (PFENet + TA, ASGNet + TA) is shown in Table \ref{tab:mIouNEW}. The forest class IoU for these two models are shown in Table \ref{tab:ForestNew}. It can be seen that the inclusion of the texture attention module in the existing models improves the accuracy of the segmentation algorithm. This improved performance further substantiates the importance of texture information in identifying forests across regions.

\begin{table}[]
\caption{Comparing the performance of the proposed method with existing few shot semantic segmentation methods (PANet \cite{wang2019panet}, ASGNet \cite{ASGNet2021}, PFENet \cite{PFENetTian2020}) in terms of overall mIoU. Also, shown are mIoU obtained for different ablation studies (i.e. evaluating performance of Support to Query (S2Q), Query to Support (Q2S) and Texture Attention (TA) modules). }
\begin{tabular}{|c|cc|cl|}
\hline
\multirow{2}{*}{method/mIoU} & \multicolumn{2}{c|}{\textbf{1 way}}                    & \multicolumn{2}{c|}{\textbf{2 way}}                                         \\ \cline{2-5} 
                             & \multicolumn{1}{c|}{\textbf{1 shot}} & \textbf{5 shot} & \multicolumn{1}{c|}{\textbf{1 shot}} & \multicolumn{1}{c|}{\textbf{5 shot}} \\ \hline
S2Q                          & \multicolumn{1}{c|}{0.121}           & 0.145           & \multicolumn{1}{c|}{0.137}           & 0.145                                \\ \hline
S2Q-TA                       & \multicolumn{1}{c|}{0.348}           & 0.349           & \multicolumn{1}{c|}{0.364}           & 0.407                                \\ \hline
S2Q + Q2S - TA               & \multicolumn{1}{c|}{0.330}           & 0.371           & \multicolumn{1}{c|}{0.334}           & 0.388                                \\ \hline
S2Q - TA + Q2S - TA          & \multicolumn{1}{c|}{0.448}           & 0.496           & \multicolumn{1}{c|}{0.451}           & 0.500                                \\ \hline
PANet (S2Q + Q2S)            & \multicolumn{1}{c|}{0.317}           & 0.364           & \multicolumn{1}{c|}{0.320}           & 0.370                                \\ \hline
ASGNet                       & \multicolumn{1}{c|}{0.181}           & 0.238           & \multicolumn{1}{c|}{0.211}           & 0.239                                \\ \hline
PFENet                       & \multicolumn{1}{c|}{0.260}           & 0.280           & \multicolumn{1}{c|}{0.248}           & 0.321                                \\ \hline
ASGNet-TA                    & \multicolumn{1}{c|}{0.221}           & 0.252           & \multicolumn{1}{c|}{0.261}           & 0.251                                \\ \hline
PFENet-TA                    & \multicolumn{1}{c|}{0.303}           & 0.333           & \multicolumn{1}{c|}{0.296}           & 0.405                                \\ \hline
S2Q -TA + Q2S (Ours)         & \multicolumn{1}{c|}{0.420}           & 0.471           & \multicolumn{1}{c|}{0.431}           & 0.473                                \\ \hline
\end{tabular}
\label{tab:mIouNEW}
\end{table}

\begin{table}[]
\caption{Comparing the performance of the proposed method with existing few show semantic segmentation methods (PANet \cite{wang2019panet}, ASGNet \cite{ASGNet2021}, PFENet \cite{PFENetTian2020}) in terms of forest class IoU. Also, shown is the forest class IoU for the ablation studies (i.e. evaluating performance of Support to Query (S2Q), Query to Support (Q2S) and Texture Attention (TA) modules). }
\label{tab:ForestNew}
\begin{tabular}{|c|cc|cl|}
\hline
\multirow{2}{*}{method/IoU (forest)} & \multicolumn{2}{c|}{\textbf{1 way}}                    & \multicolumn{2}{c|}{\textbf{2 way}}                                         \\ \cline{2-5} 
                                     & \multicolumn{1}{c|}{\textbf{1 shot}} & \textbf{5 shot} & \multicolumn{1}{c|}{\textbf{1 shot}} & \multicolumn{1}{c|}{\textbf{5 shot}} \\ \hline
S2Q                                  & \multicolumn{1}{c|}{0.473}           & 0.539           & \multicolumn{1}{c|}{0.496}           & 0.591                                \\ \hline
S2Q-TA                               & \multicolumn{1}{c|}{0.517}           & 0.556           & \multicolumn{1}{c|}{0.534}           & 0.618                                \\ \hline
S2Q + Q2S - TA                       & \multicolumn{1}{c|}{0.510}           & 0.609           & \multicolumn{1}{c|}{0.526}           & 0.641                                \\ \hline
S2Q - TA + Q2S - TA                  & \multicolumn{1}{c|}{0.633}           & 0.691           & \multicolumn{1}{c|}{0.621}           & 0.705                                \\ \hline
PANet (S2Q + Q2S)                    & \multicolumn{1}{c|}{0.465}           & 0.521           & \multicolumn{1}{c|}{0.501}           & 0.580                                \\ \hline
ASGNet                               & \multicolumn{1}{c|}{0.181}                & 0.238                 & \multicolumn{1}{c|}{0.211}                & 0.239                                      \\ \hline
PFENet                               & \multicolumn{1}{c|}{0.260}                & 0.280                 & \multicolumn{1}{c|}{0.248}                & 0.321                                      \\ \hline
ASGNet-TA                            & \multicolumn{1}{c|}{0.221}                & 0.252                 & \multicolumn{1}{c|}{0.261}                & 0.251                                      \\ \hline
PFENet-TA                            & \multicolumn{1}{c|}{0.303}                & 0.333                 & \multicolumn{1}{c|}{0.296}                & 0.405                                      \\ \hline
S2Q -TA + Q2S (Ours)                 & \multicolumn{1}{c|}{0.626}           & 0.678           & \multicolumn{1}{c|}{0.624}           & 0.699                                \\ \hline
\end{tabular}
\end{table}

\section{Conclusion and Future work}

The satellite-based forest monitoring system provides a cost-effective tool to identify the change in forest coverage of a region. The existing forest identification methods are based on supervised learning approaches that depend on manually annotated data and are limited to a particular geographical region. This work proposes a texture-based prototype alignment few-shot semantic segmentation network for forest identification. In this work, support prototype features are enhanced by incorporating texture features to highlight the difference in the texture of forest and other classes. In addition, the proposed method focuses on cross-geography generalization wherein the model is trained on the images of South Asia and evaluated on images of Central Europe. The proposed method is compared with the existing few shot segmentation methods and an IoU of 0.62 (1-way 1-shot) is observed for forest class as compared to existing methods (0.46 for PANet \cite{wang2019panet}). Interestingly, the proposed method can accurately differentiate between forest and other classes (cropland, water, etc) compared to PANet. The results demonstrate that features learned for identifying tropical forests can be transferred to identify temperate forests with the help of very few labeled images (one image for 1-shot) of temperate forests. 
In future, this study would be extended to identify the forest of other geographical regions (South America, Russia, Canada, China), thus creating an opportunity to develop a global forest identification tool.

	\bibliographystyle{ieee_fullname}
	\bibliography{NeurIPS-bib}

\end{document}